\documentclass[runningheads]{llncs}

\usepackage{graphicx}
\usepackage[square,sort,comma,numbers,sectionbib]{natbib}
\usepackage[small]{caption}
\usepackage{subcaption}
\usepackage{xcolor}
\usepackage{bbm}
\usepackage{amssymb}
\usepackage{amsmath}
\usepackage{booktabs}
\usepackage{algorithm}
\usepackage{algorithmic}
\usepackage{array}
\usepackage{url}

\newcommand{\cut}[1]{}

\newcolumntype{H}{>{\setbox0=\hbox\bgroup}c<{\egroup}@{}}
\begin{document}
\title{Use the Detection Transformer as a\\Data Augmenter}
%
%
\author{Luping Wang \and
Bin Liu\thanks{Corresponding author}}
\authorrunning{L. Wang and B. Liu}

\institute{Research Center for Applied Mathematics and Machine Intelligence, \\Zhejiang Lab, Hangzhou 311121, China\\
\email{\{wangluping,liubin\}@zhejianglab.com}\\
}
\maketitle              
\begin{abstract}
Detection Transformer (DETR) is a Transformer architecture based object detection model. In this paper, we demonstrate that it can also be used as a data augmenter. We term our approach as DETR assisted CutMix, or DeMix for short. DeMix builds on CutMix, a simple yet highly effective data augmentation technique that has gained popularity in recent years. CutMix improves model performance by cutting and pasting a patch from one image onto another, yielding a new image. The corresponding label for this new example is specified as the weighted average of the original labels, where the weight is proportional to the area of the patch. CutMix selects a random patch to be cut. In contrast, DeMix elaborately selects a semantically rich patch, located by a pre-trained DETR. The label of the new image is specified in the same way as in CutMix. Experimental results on benchmark datasets for image classification demonstrate that DeMix significantly outperforms prior art data augmentation methods including CutMix. Oue code is available at \url{https://github.com/ZJLAB-AMMI/DeMix}.
\end{abstract}
\section{Introduction}
\label{introduction}
Data augmentation is a technique used in machine learning where existing data is modified or transformed to create new data.
The principle behind data augmentation is that even small changes to existing data can create new useful examples for training. For example, flipping an image horizontally can create a new training example that is still representative of the original object.
In general, the goal of data augmentation is to increase the diversity of the training data while preserving its underlying structure. By introducing variations into the training data, models can learn to generalize better and perform well on unseen data.
Data augmentation has been around for decades, but recent advances in deep learning have made it more widely used and effective. It has been commonly used in computer vision tasks like image classification~\cite{mikolajczyk2018data,perez2017effectiveness,fawzi2016adaptive}, object detection~\cite{zoph2020learning,zhong2020random,montserrat2017training}, and segmentation~\cite{chaitanya2021semi,xu2020automatic,ghiasi2021simple,zhao2019data}, as well as in natural language processing~\cite{chen2021hiddencut,kafle2017data,bayer2023data}.
As a powerful technique that can significantly improve model performance with little additional effort or cost, its widespread adoption and continued development are indicative of its importance in modern machine learning.

There are many typical methods of data augmentation, including flipping, rotating, scaling, cropping, adding noise, and changing color, which can be applied randomly or according to specific rules, depending on the task and the desired results~\cite{shorten2019survey}. Recent years also witnessed notable developments in data augmentation techniques, among which CutMix~\cite{cutmix}, a method that combines parts of multiple images to create new training samples, is our focus in this work.

CutMix is a simple yet highly effective data augmentation technique that has gained popularity in recent years. Given a pair of training examples $A$ and $B$ and their associated labels $y_A$ and $y_B$, CutMix selects a random crop region of $A$ and replaces it with a patch of the same size cut from $B$, yielding a new data example. The corresponding label for this new example is specified as the weighted average of the original labels, where the weight is proportional to the area of the patch. As the patch to be cut is randomly selected, it may totally come from the background or the area of the object or an area that mixes the background and the object, while the contribution of this patch to the label of the resulting new image is deterministic. We argue that this is not reasonable. For example, if the patch selected to be cut is all from the background, then its contribution to the label of the resulting new image should be negligible, while using CutMix, its contribution is proportional to its area.

Motivated by the aforementioned flaw of CutMix, we propose a knowledge-guided CutMix, where the knowledge comes from a pre-trained detection transformer (DETR) model. We term our approach DETR assisted CutMix, or DeMix for short. DETR is an object detection model based on the Transformer architecture~\cite{detr,han2022survey,parmar2018image}. Unlike traditional object detection models, DETR directly models the object detection task as a set matching problem, and uses a Transformer encoder to process input images and a decoder to generate object sets, achieving end-to-end object detection. DeMix takes advantage of the following desirable properties of DETR
\begin{itemize}
  \item Given an image input to a pre-trained DETR, it can provide an estimate of the class, bounding box position, and corresponding confidence score for each object involved in this image;
  \item It can detect objects of different numbers and sizes simultaneously.
\end{itemize}

As the pre-trained DETR model is trained based on datasets that are different from our target dataset, the class labels it provides can be meaningless, while bounding box positions it provides are surprisingly informative for us to use in DeMix. DeMix cuts the image patch associated with one of the bounding boxes, given by DETR, in an image example, then resizes and pastes it onto a random crop region of another example, to create the new example. The label of this new example is specified in the same way as CutMix. 

In principle, DeMix provides an elaborate improvement to CutMix by borrowing knowledge from a pre-trained DETR. Even if the knowledge borrowed is totally inaccurate or meaningless, then DeMix reduces to CutMix. Our major contributions can be summarized as follows
\begin{itemize}
  \item We demonstrates how DETR can be used as a tool for data augmentation, resulting in a new approach DeMix;
  \item We evaluate the performance of DeMix on several different fine-grained image classification datasets. Experimental results demonstrate that DeMix significantly outperforms all competitor methods, including CutMix;
  \item Our work sheds light on how to improve a general-purpose technique (data augmentation here) using a special-purpose pre-trained model (DETR here) without fine-tuning.
\end{itemize}
%

\section{Related Works}
\label{background}
In this section, we briefly introduce DETR followed by related works on data augmentation techniques, especially CutMix, which are closely related to our DeMix.
\subsection{DETR}
\begin{figure}[htbp]
    \centering
    \includegraphics[width=\textwidth]{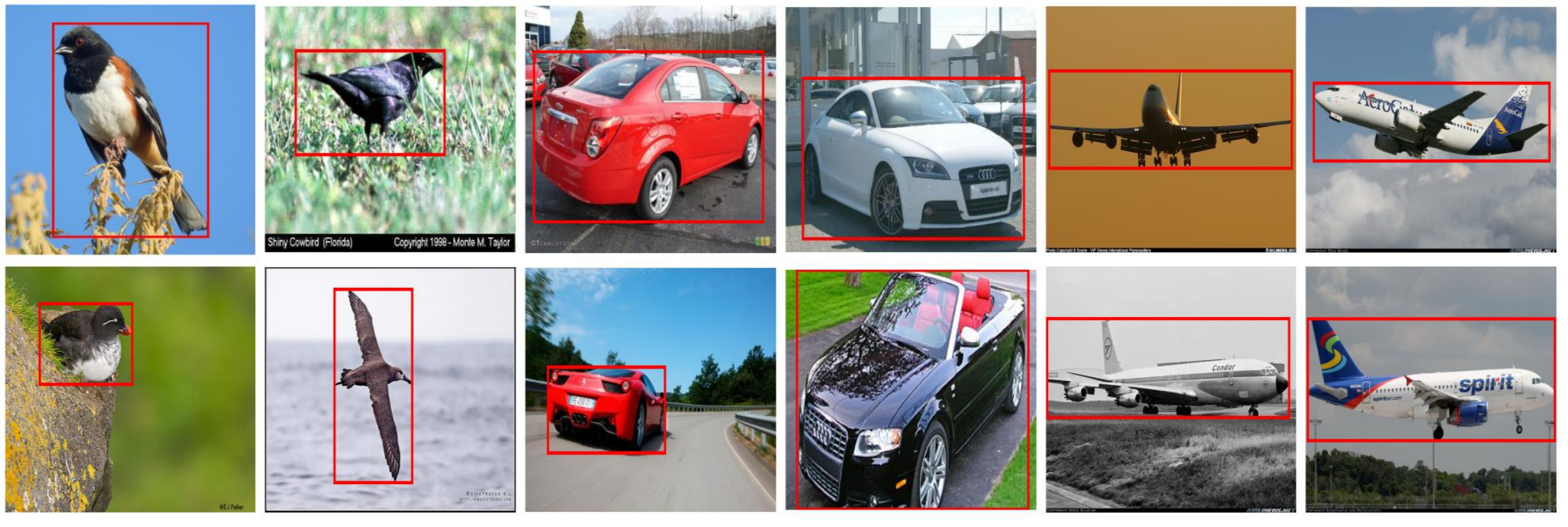}
    \caption{Object detection with DETR}
    \label{fig:detr_res}
\end{figure}

Detection Transformer (DETR) is a type of object detection model based on the Transformer architecture~\cite{detr,han2022survey,parmar2018image}. Unlike traditional object detection models, DETR directly models the object detection task as a set matching problem, and uses a Transformer encoder to process input images and a decoder to generate object sets, achieving end-to-end object detection~\cite{detr}.
In DETR, the image is divided into a set of small patches and each patch is mapped to a feature vector using a convolutional neural network (CNN) encoder. Then, the output of the encoder is passed as input to the Transformer encoder to allow interactions among the features in both spatial and channel dimensions. Finally, the decoder converts the output of the Transformer encoder into an object set that specifies the class, bounding box position, and corresponding confidence score for each object. DETR does not require predefined operations such as non-maximum suppression or anchor boxes. It can detect objects of different numbers and sizes simultaneously. Fig.~\ref{fig:detr_res} shows detection results given by a pre-trained DETR model on some image examples in the dataset used in our experiment.

DETR is developed for object detection, while, in this paper, we demonstrate that it could also be used as a data augmenter. Specifically, we leverage DETR to locate a semantically rich patch associated with an object for use in CutMix.
\subsection{Data Augmentation}
Data augmentation is a technique used in machine learning to increase the size of a training dataset by generating new examples from existing ones. The goal of data augmentation is to improve the generalization ability of machine learning models by introducing more variations in the training data.

There are many existing works on data augmentation, which can be broadly categorized into four groups as follows.
\begin{enumerate}
  \item Traditional methods: These include commonly used techniques such as random cropping, resizing, flipping, rotating, color jittering, and adding noise~\cite{shorten2019survey}. The strength of traditional methods is that they are simple and easy to implement, and can effectively generate new samples from existing ones. However, they may not always work well for complex datasets or tasks, as they do not account for higher-level features and structures.
  \item Generative methods: These use generative models, such as variational autoencoders (VAEs)  and generative adversarial networks (GANs), to synthesize new data samples \cite{wu2019data,yang2021ida,chadebec2022data}. The strength of generative methods is that they can generate high-quality and diverse samples that are similar to the real data distribution, which can help improve model generalization. However, they require large amounts of computational resources and training data to build the generative models, which can be a limitation in some cases.
  \item Adversarial methods: These use adversarial attacks to perturb the existing data samples to generate new ones \cite{zhao2020maximum,zhang2019dada,peng2018jointly}. The strength of adversarial methods is that they can generate realistic and diverse augmented data samples, which can help improve model robustness. However, they require careful design and tuning of hyperparameters to avoid overfitting.
  \item Methods that create new training examples by mixing parts of multiple images together such as Mixup~\cite{mixup} and CutMix~\cite{cutmix}. Mixup linearly combines data from different examples, while CutMix cuts and pastes patches of images. These methods are much simpler to implement than generative and adversarial methods, while can achieve much better performance than traditional methods.
\end{enumerate}
In summary, each data augmentation method has its own strengths and limitations.

DeMix proposed here is an algorithmic improvement to the aforementioned method CutMix, which has gained popularity in recent years, since it is simple to implement, while can achieve much better performance than traditional methods. In this paper, we propose DeMix, which is as simple as CutMix, while outperforms it significantly.

In CutMix, the image patch to be cut is randomly selected, while the contribution of this patch to the label of the resulting new image is deterministically proportional to its area. As aforementioned, this is not reasonable. For example, if the patch to be cut is all from the background, then its contribution to the label of the resulting new image should be negligible, while using CutMix, it can be large.

Several work has been proposed to address this flaw of CutMix, where the basic idea is to select a semantically rich instead of a random patch to be cut and paste, see e.g., SaliencyMix~\cite{saliencymix}, the class activation map (CAM) based method~\cite{zhang2021new}, Keepaugment~\cite{gong2021keepaugment}, and SnapMix~\cite{snapmix}. All these advanced methods require image pre-processing, such as computing the saliency map or CAM, prior to data augmentation, while, in contrast, DeMix does not need any pre-processing operation before generating new image examples.
\section{DeMix: DETR Assisted CutMix}
\label{methods}
\begin{figure}[htbp]
    \centering
    \includegraphics[width=\textwidth]{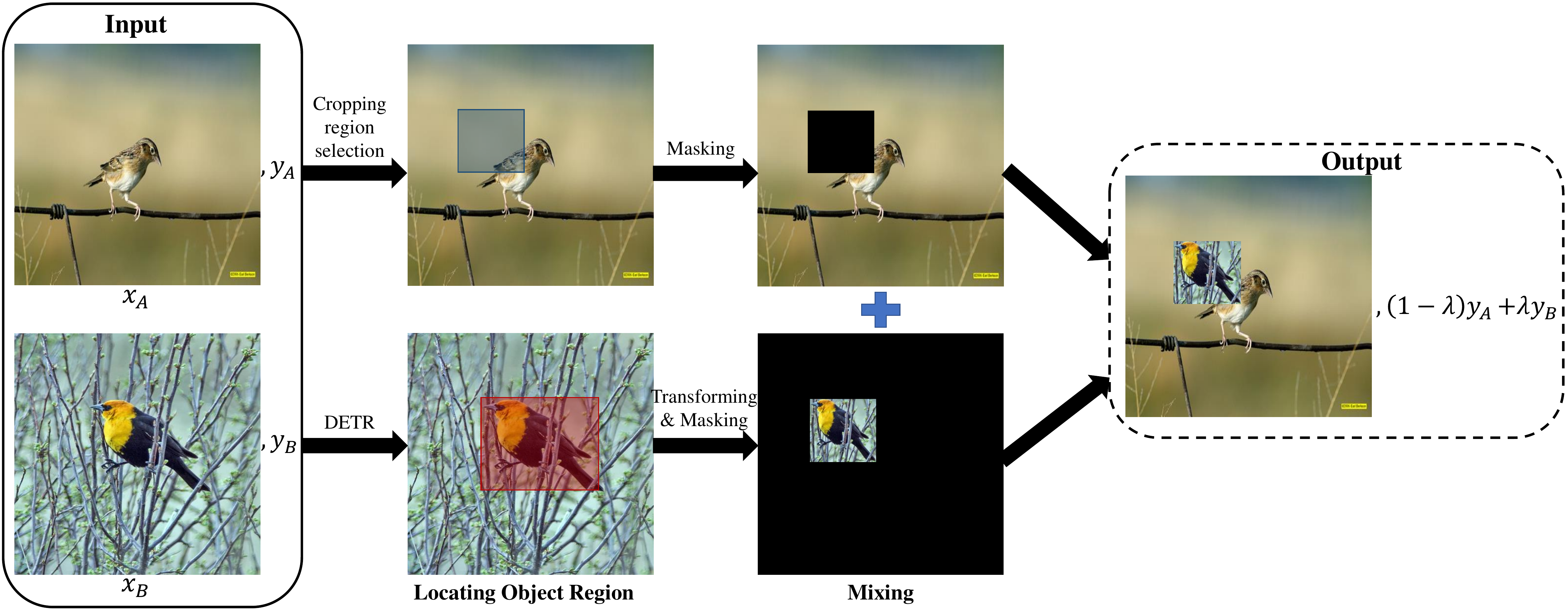}
    \caption{An example show of the operations of DeMix. Given a pair of a target image and a source image, denoted as $x_A$ and $x_B$, with one-hot labels $y_A$ and $y_B$, respectively, DeMix starts by randomly selecting a cropping region for $x_A$, and locating the object region for $x_B$ based on the object bounding box outputted by a pre-trained DETR. The `Transforming' operation performed on $x_B$ resizes and relocates the image patch located by DETR to make its size and position consistent with those of the cropping region in $x_A$.}
    \label{fig:demix}
\end{figure}

In this section, we describe our DeMix method in detail. Given a pair of image examples, DeMix uses two operations to generate a new image. The first operation employs a pre-trained DETR to identify bounding box positions for each object in the source image denoted as $x_B$. The second operation cuts an image patch associated with one bounding box, resize it, and then pastes it onto a randomly selected crop region of the other image, termed target image and denoted as $x_A$, yielding the new image example. The label of this new image is a weighted average of labels of the original images. Fig.~\ref{fig:demix} illustrates the operations that make up DeMix.

In mathematical terms, the process of generating a new image can be explained as follows.:
\begin{equation}\label{eq:demix}
    \begin{array}{ccl}
         \tilde{x} &=&  (\bold{1 - M_{\lambda}}) \odot x_{A} + T(\bold{M_{B}} \odot x_{B})\\
         \tilde{y} &=&  (1 - \lambda) y_{A} + \lambda y_{B}
    \end{array}
\end{equation}
where $y_A$ and $y_B$ denote labels of $x_A$ and $x_B$, respectively, in the form of one-hot vectors, $\bold{M_{\lambda}}$ and $\bold{M_{B}}$ are binary mask matrices with dimensions $W \times H$, $W$ and $H$ denote the width and height of the images, $0<\lambda<1$ is a hyper-parameter defined as the ratio of the area of the randomly selected crop region of $x_A$ to the full area of $x_A$, $\bold{M_{\lambda}}$ denotes the binary mask matrix that defines the aforementioned crop region, $\bold{M_{B}}$ is the binary mask matrix associated to the object bounding box given by DETR, $\odot$ denotes the element-wise multiplication operator, $\bold{1}$ represents a matrix of an appropriate size whose elements are all 1, and finally $T(\cdot)$ denotes a linear transformation that aligns the size and position of the cut patch to be consistent with those of the crop region in $x_A$ on which this patch will be pasted.
\subsection{Discussions on the algorithm design of DeMix}
In DeMix, we select the target and source images, which are used for generating a new image, in the same way as in CutMix, namely, they are randomly selected from the training set. Since we use a pre-trained DETR for DeMix, which means that we do not need to train the DETR model, the computational overhead of DeMix is comparable to CutMix.

DeMix is based on CutMix, with the same ``random cropping" and linear label generation operations. The concept of random cropping has been widely used in deep learning (DL) data augmentation techniques like CutMix~\cite{cutmix} and Cutout~\cite{devries2017improved}. The Dropout approach, which is often used for DL regularization \cite{baldi2013understanding,srivastava2014dropout}, is also a form of ``random cropping" in essence. However, instead of image patches, it crops neural network weights. Empirically speaking, ``random cropping" is a simple yet effective strategy for DL regularization. Its basic mechanism is that DL generally follows ``shortcut" learning, making predictions based on ``shortcut" features embedded in the training dataset \cite{geirhos2020shortcut}. For instance, if all cows in the training set appear with grass, the DL model could link grass features to cow existence. If a cow appears on a beach without grass in a test image, the model will predict that there is not a cow. By utilizing ``random cropping," DL lessens its reliance on shortcut features, reducing the overfitting probability and enhancing the model's generalization ability.

\begin{figure}[htbp]
    \centering
    \includegraphics[width=\textwidth]{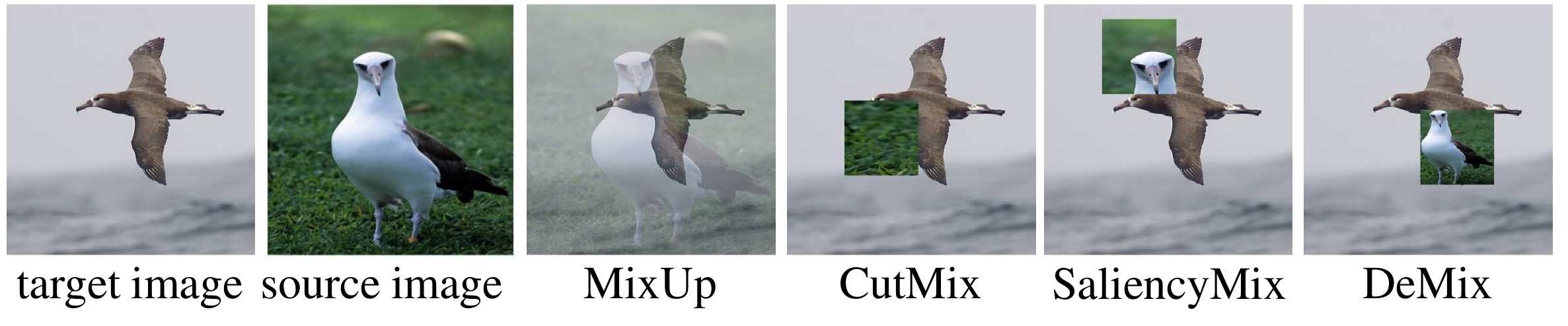}
    \caption{Comparison between related data augmentation techniques}
    \label{fig:mix_compare}
\end{figure}

We display data augmentation results of MixUp~\cite{mixup}, CutMix~\cite{cutmix}, SaliencyMix~\cite{saliencymix}, and DeMix in Fig.~\ref{fig:mix_compare}. As is shown, DeMix cuts and resizes a patch that covers the whole object region from the source image, while SaliencyMix selects a patch corresponding to the most salient box region that only covers a part of the object. CutMix randomly selects a patch to be cut, which may only contain the background region. MixUp mixes the target and source images through linear combination, which may lead to local ambiguity and unnaturalness in the generated images, as addressed by~\cite{cutmix}.

As DeMix is a pre-trained DETR assisted data augmentation technique, its performance is strongly connected to the quality of the DETR model. When the DETR model works well to accurately locate the object region in the source image, then DeMix could succeed in selecting a semantically rich patch to be cut and mixed. When the DETR model being used fails to accurately locate the object region in the source image, then, in principle, DeMix reduces to CutMix, since the patch to be cut can be seen as one randomly selected.
\section{Experiments}
In this section, we evaluate the performance of our DeMix method through experiments on image classification tasks that involve different datasets, different neural network architectures of different sizes. Related modern data augmentation techniques, see subsection \ref{sec:methods}, are used for performance comparison.
\subsection{Experimental Setting}
\subsubsection{Datasets}
In our experiments, we selected three benchmark fine-grained image datasets for use, namely CUB-200-2011~\cite{cub}, Stanford-Cars~\cite{cars}, and FGVC-Aircraft~\cite{aircraft}. For simplicity, we refer to these three datasets as CUB, Cars, and Aircraft, respectively, in what follows.
\subsubsection{Network Architectures}
In order to perform a comprehensive evaluation on the performance of our method, we selected 6 different network architectures in our experiments, including ResNet-(18, 34, 50, 101)~\cite{resnet}, InceptionV3~\cite{inception}, and DenseNet121~\cite{densenet}.
\subsubsection{Comparison Methods}\label{sec:methods}
For performance comparison, we selected modern data augmentation techniques in our experiments including CutMix~\cite{cutmix}, SaliencyMix~\cite{saliencymix}, MixUp~\cite{mixup}, and CutOut~\cite{cutout}. We also include a baseline method that refers to a model trained without using any data augmentation technique.
\subsubsection{Futher details on model training}
We used the open-source pre-trained DETR model $detr\_resnet50$ included in the Pytorch-torchvision package. The initial feature extractor parameter values are set to be equal with those of the ResNet50 feature extractor pre-trained on the ImageNet-1K dataset~\cite{he2016deep}, and the entire DETR model is trained based on the MS-COCO dataset~\cite{lin2014microsoft}. Fig.~\ref{fig:detr_res} demonstrates the detection performance of the aforementioned DETR model on some image examples involved in our experiments.

In our image classification tasks, we followed~\cite{snapmix} to specify hyper-parameter values for model training. Specifically, we chose the stochastic gradient descent (SGD) with momentum as the optimizer, and set the momentum factor at 0.9. The initial learning rate for the pre-trained weights was set to 0.001, while that for other weights was set to 0.01. If training from scratch, the initial learning rate for all trainable weights was set to 0.01. When using pre-trained weights, the model was trained for 200 epochs and decayed the learning rate by factor 0.1 at 80, 150, and 180 epoch; otherwise, the model was trained for 300 epochs and decayed the learning rate by factor 0.1 at 150, 225, and 270 epoch.
\subsection{Experimental Results}
We used different data augmentation techniques in image classification and evaluated the performance of each data augmentation technique via its associated classification accuracy. In Tables ~\ref{tab:pt1_res18_34} and ~\ref{tab:pt1_res50_101}, we show the image classification performance, in terms of the average top-1 accuracy, with respect to ResNet architectures with different depths. Results of the baseline method, MixUp, CutOut, and CutMix are directly quoted from~\cite{snapmix}. The training setting for the other methods, namely SaliencyMix and DeMix, were set as the same as \cite{snapmix} to guarantee that the performance comparison is fair.

We see that DeMix achieved the best performance in almost all the experiments compared to other methods.
It also shows that SaliencyMix does not provide a significant performance gain over CutMix on these datasets, while DeMix does. We argue that it is because the discriminative parts of an image are not located in the salient region captured by SaliencyMix, while they are located in the object region detected by the DETR model employed by DeMix.

Furthermore, we found that DeMix performs more stable, compared to other methods, when the network depth varies. For example, on the dataset CUB, CutMix and SaliencyMix perform poorly with shallower network architectures ResNet18, while show significant performance improvement with deeper architectures like ResNet101. This may be because the image samples generated by CutMix and SaliencyMix are more noisy than those generated by DeMix, and deeper networks are better at handling noisy samples. Overall, regardless of the network depth, DeMix outperforms the other methods significantly.
\begin{table}[htbp]
  \centering
  \caption{Top-1 accuracy (\%) of each method for image classification tasks on datasets CUB, Cars, and Aircraft. The classification network is initialized by a pre-trained ResNet18 or ResNet34. The best performance is marked in bold.}
    \begin{tabular}{l|cc|cc|cc}
    \toprule
          & \multicolumn{2}{c|}{\textbf{CUB}} & \multicolumn{2}{c|}{\textbf{Cars}} & \multicolumn{2}{c}{\textbf{Aircraft}} \\
          & ResNet18 & ResNet34 & ResNet18 & ResNet34 & ResNet18 & ResNet34 \\
    \midrule
    baseline & 82.35 & 84.98 & 91.15 & 92.02 & 87.80 & 89.92 \\
    MixUp & \textbf{83.17} & 85.22 & 91.57 & 93.28 & 89.82 & 91.02 \\
    CutOut & 80.54 & 83.36 & 91.83 & 92.84 & 88.58 & 89.90 \\
    CutMix & 80.16 & 85.69 & 92.65 & 93.61 & 89.44 & 91.26 \\
    SaliencyMix & 80.69 & 85.17 & 93.17 & 93.94 & \textbf{90.61} & 91.72 \\
    DeMix & 82.86 & \textbf{86.69} & \textbf{93.37} & \textbf{94.49} & 90.52 & \textbf{93.10} \\
    \bottomrule
    \end{tabular}%
  \label{tab:pt1_res18_34}%
\end{table}
\begin{table}[htbp]
  \centering
  \caption{Top-1 accuracy (\%) of each method for image classification tasks on datasets CUB, Cars, and Aircraft. The classification network is initialized by a pre-trained ResNet50 or ResNet101. The best performance is marked in bold.}
    \begin{tabular}{l|cc|cc|cc}
    \toprule
          & \multicolumn{2}{c|}{\textbf{CUB}} & \multicolumn{2}{c|}{\textbf{Cars}} & \multicolumn{2}{c}{\textbf{Aircraft}} \\
          & ResNet50 & ResNet101 & ResNet50 & ResNet101 & ResNet50 & ResNet101 \\
    \midrule
    baseline & 85.49 & 85.62 & 93.04 & 93.09 & 91.07 & 91.59 \\
    MixUp & 86.23 & 87.72 & 93.96 & 94.22 & 92.24 & 92.89 \\
    CutOut & 83.55 & 84.70 & 93.76 & 94.16 & 91.23 & 91.79 \\
    CutMix & 86.15 & 87.92 & 94.18 & 94.27 & 92.23 & 92.29 \\
    SaliencyMix & 86.35 & 87.59 & 94.23 & 94.22 & 92.41 & 92.77 \\
    DeMix & \textbf{86.93} & \textbf{88.23} & \textbf{94.59} & \textbf{94.81} & \textbf{93.76} & \textbf{94.27} \\
    \bottomrule
    \end{tabular}%
  \label{tab:pt1_res50_101}%
\end{table}

We further conducted experiments to evaluate the performance of DeMix on other neural network architectures, namely InceptionV3 and DenseNet121. The results are shown in table ~\ref{tab:pt1_iv3_d121}. Again, we see a significant performance improvement given by DeMix over the other methods.
\begin{table}[htbp]
  \centering
  \caption{Top-1 accuracy (\%) of each method for image classification tasks on datasets CUB, Cars, and Aircraft. The classification network is initialized by a pre-trained InceptionV3 or DenseNet121. The best performance is marked in bold.}
    \begin{tabular}{l|cc|cc|cc}
    \toprule
          & \multicolumn{2}{c|}{\textbf{CUB}} & \multicolumn{2}{c|}{\textbf{Cars}} & \multicolumn{2}{c}{\textbf{Aircraft}} \\
          & InceptionV3 & DenseNet121 & InceptionV3 & DenseNet121 & InceptionV3 & DenseNet121 \\
    \midrule
    baseline & 82.22 & 84.23 & 93.22 & 93.16 & 91.81 & 92.08 \\
    MixUp & 83.83 & 86.65 & 92.23 & 93.21 & 92.02 & 91.42 \\
    CutMix & 84.31 & 86.11 & 93.94 & 94.25 & 92.71 & 93.40 \\
    SaliencyMix & 85.07 & 85.26 & \textbf{94.18} & 93.65 & 93.58 & 92.95 \\
    DeMix & \textbf{85.12} & \textbf{87.38} & 94.13 & \textbf{94.29} & \textbf{93.85} & \textbf{94.27} \\
    \bottomrule
    \end{tabular}%
  \label{tab:pt1_iv3_d121}%
\end{table}%

In previous experiments, we fine-tuned pre-trained classification models using augmented training datasets in the modeling training phase.
We also conducted experiments wherein we train the classification models from scratch. In this way, we get a clearer performance evaluation, since it avoids the impact of pre-training on data augmentation performance evaluation. The corresponding results are shown in table ~\ref{tab:pt0_res18_50}. We see that DeMix performs best in most cases and comparably in the other ones. In particular, on the CUB dataset, DeMix gives a significant performance improvement compared to CutMix and SaliencyMix.
It may be because that, in the CUB dataset, images of different classes have more subtle differences among each other, thus requiring training samples of higher quality; and DeMix can generate samples of higher quality than CutMix and SaliencyMix.
\begin{table}[htbp]
  \centering
  \caption{Top-1 accuracy (\%) of each method for image classification tasks on datasets CUB, Cars, and Aircraft. The classification network architecture is set as ResNet18 or ResNet50, the same as in Table 1, while here the model is trained from scratch, other than pre-trained as shown in Table 1. The best performance is marked in bold.}
    \begin{tabular}{l|cc|cc|cc}
    \toprule
          & \multicolumn{2}{c|}{\textbf{CUB}} & \multicolumn{2}{c|}{\textbf{Cars}} & \multicolumn{2}{c}{\textbf{Aircraft}} \\
          & ResNet18 & ResNet50 & ResNet18 & ResNet50 & ResNet18 & ResNet50 \\
    \midrule
    baseline & 64.98 & 66.92 & 85.23 & 84.63 & 82.75 & 84.49 \\
    MixUp & 67.63 & \textbf{72.39} & 89.14 & 89.69 & 86.38 & 86.59 \\
    CutMix & 60.03 & 65.28 & 89.11 & 90.13 & 85.60 & 86.95 \\
    SaliencyMix & 65.60 & 67.03 & 88.53 & 89.81 & 86.95 & \textbf{88.81} \\
    DeMix & \textbf{70.00} & 71.80 & \textbf{89.83} & \textbf{91.72} & \textbf{88.48} & 88.66 \\
    \bottomrule
    \end{tabular}%
  \label{tab:pt0_res18_50}%
\end{table}%

\begin{figure}[htbp]
    \centering
    \includegraphics[width=\textwidth]{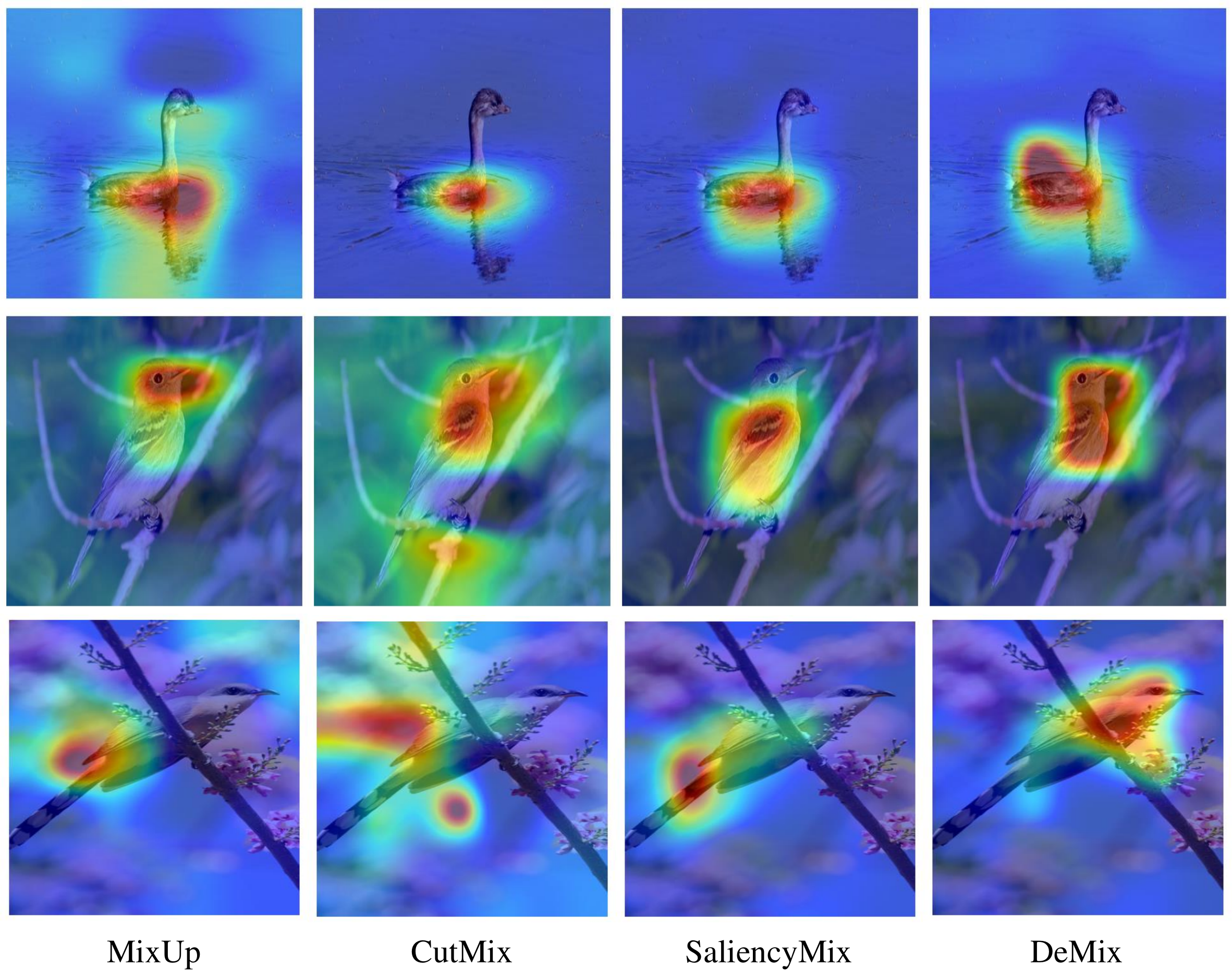}
    \caption{Visualizations of class activation mapping (CAM) on 3 test image examples}
    \label{fig:cam_analysis}
\end{figure}

In order to understand why DeMix performs better than the other methods involved in our experiments, we investigate the class activation mapping (CAM) associated with each data augmentation technique. CAM is a technique used in deep learning based computer vision to visualize the regions of an image that are most important for a neural network's classification decision. CAM generates a heatmap that highlights the regions of the image that contributed most to the predicted output class, allowing humans to better understand how the model is making its predictions. We checked CAMs given by the classification models trained with aid of different data augmentation techniques. The visualization results on 3 test image examples are shown in Fig.~\ref{fig:cam_analysis}.
As is shown, using DeMix, the regions of the image that contributed most to the predicted output class match the real objects' regions to a greater extent than using the other data augmentation techniques. For example, for the 2nd test image, the classification model trained with aid of MixUp mainly uses the head of the bird, the model associated with SaliencyMix mainly uses the body of the bird, while the model corresponding to DeMix uses the head and a part of the body together, to generate the predicted label. For the 3rd test image, it is clearer that the model associated with DeMix selects a more appropriate region for use in making class predictions.
%
\subsection{Further Experiments}
We conduct an experiment to investigate the influence of the hyperparameter $\lambda$, namely the ratio of the area of the randomly selected crop region, on performance of DeMix. See the result in Table \ref{tab:lam_analysis}, which shows that the performance of DeMix is not very sensitive to the value of $\lambda$. Note that our DeMix is built upon CutMix. It utilizes a random $\lambda$, the same as CutMix, to enhance sample diversity in the augmented dataset, which has been demonstrated to be beneficial for improving the model's performance in terms of generalization.
\begin{table}[htbp]
  \centering
  \caption{Influence of $\lambda$ on performance of DeMix on the image classification task using the CUB dataset}
    \begin{tabular}{cccccccccc}
    \toprule
       $\lambda$   & 0.1   & 0.2   & 0.3   & 0.4   & 0.5   & 0.6   & 0.7   & 0.8   & 0.9 \\
    \midrule
    ResNet18 & 83.83 & 83.28 & 82.93 & 82.78 & 82.67 & 82.36 & 82.02 & 81.43 & 80.83 \\
    ResNet50 & 87.33 & 87.02 & 86.90 & 86.90 & 86.87 & 86.61 & 86.49 & 85.90 & 85.40 \\
    \bottomrule
    \end{tabular}%
  \label{tab:lam_analysis}%
\end{table}

We also consider long-tailed recognition tasks on the CUB dataset. Performance comparison results between DeMix with CutMix and SaliencyMix with different imbalance ratios are presented in Table \ref{tab:cub_longtailed}. It is shown that, for both architectures ResNet18 and ResNet50, DeMix performs best.
\begin{table}[htbp]
  \centering
  \caption{Top-1 accuracy (\%) comparison on long-tailed CUB dataset with different imbalance ratios}
    \begin{tabular}{l|cc|cc}
    \toprule
          & \multicolumn{2}{c|}{ResNet18} & \multicolumn{2}{c}{ResNet50} \\
    Imbalance Ratio &  50\% & 10\% &  50\% & 10\% \\
    \midrule
    CutMix &  32.15 & 52.50 &  38.54 & 62.63 \\
    SaliencyMix &  29.58 & 45.53 &  33.76 & 54.90 \\
    DeMix &  \textbf{32.97} & \textbf{52.99} & \textbf{38.89} & \textbf{62.91} \\
    \bottomrule
    \end{tabular}%
  \label{tab:cub_longtailed}%
\end{table}%
\section{Conclusion}
In this paper, we demonstrated that a pre-trained object detection model, namely DETR, can be used as a tool for developing powerful data augmentation techniques. Specifically, we found that a DETR model pre-trained on the MS-COCO dataset~\cite{lin2014microsoft} can be used to locate semantically rich patches in images of other datasets, such as CUB-200-2011~\cite{cub}, Stanford-Cars~\cite{cars}, and FGVC-Aircraft~\cite{aircraft}. Then we proposed DeMix, a novel data augmentation technique that employs DETR to assist CutMix in locating semantically rich patches to be cut and pasted. Experimental results on several fine-grained image classification tasks that involve different network depths and different network architectures demonstrate that our DeMix performs strikingly better than prior art methods. Our work thus suggests (or confirms) that leveraging the power of a pre-trained (large) model directly, without fine-tuning, is a promising direction for future research to improve task-specific performance.
\bibliographystyle{splncs04}
\bibliography{ref}
\end{document}